%% file: paper.tex
\documentclass[]{bytedance_seed}

\usepackage[toc,page,header]{appendix}

\usepackage{makecell}

\newcommand{\bench}[1]{NL2Repo-Bench}

\title{\bench{}: Towards Long-Horizon Repository Generation Evaluation of Coding Agents}

\affiliation[1]{ByteDance Seed China}
\affiliation[2]{M-A-P}
\affiliation[3]{2077AI}
\affiliation[4]{Humanlaya Data}
\affiliation[5]{Nanjing University}
\affiliation[6]{Peking University}
\affiliation[7]{Beijing University of Posts and Telecommunications}
\affiliation[8]{Beihang University}

\contribution{Full author list in Contributions}

\abstract{

Recent advances in coding agents suggest rapid progress toward autonomous software development, yet existing benchmarks fail to rigorously evaluate the long-horizon capabilities required to build complete software systems. Most prior evaluations focus on localized code generation, scaffolded completion, or short-term repair tasks, leaving open the question of whether agents can sustain coherent reasoning, planning, and execution over the extended horizons demanded by real-world repository construction. To address this gap, we present NL2Repo Bench, a benchmark explicitly designed to evaluate the long-horizon repository generation ability of coding agents. Given only a single natural-language requirements document and an empty workspace, agents must autonomously design the architecture, manage dependencies, implement multi-module logic, and produce a fully installable Python library. 
Our experiments across state-of-the-art open- and closed-source models reveal that long-horizon repository generation remains largely unsolved: even the strongest agents achieve below 40\% average test pass rates and rarely complete an entire repository correctly. Detailed analysis uncovers fundamental long-horizon failure modes, including premature termination, loss of global coherence, fragile cross-file dependencies, and inadequate planning over hundreds of interaction steps. NL2Repo Bench establishes a rigorous, verifiable testbed for measuring sustained agentic competence and highlights long-horizon reasoning as a central bottleneck for the next generation of autonomous coding agents.

}

\date{\today}

\checkdata[Project Page]{\url{https://github.com/multimodal-art-projection/NL2RepoBench}}

\begin{document}
\maketitle


\input{sections/introduction}

\input{sections/relatedwork}

\input{sections/approach}
\input{sections/experiments}
\clearpage
\input{sections/contribution}

\clearpage

\bibliographystyle{plainnat}
\bibliography{main}

\clearpage

\beginappendix

\input{sections/appendix}

\end{document}

%% file: sections/introduction.tex
\section{Introduction}

Large language models (LLMs) have rapidly evolved from passive code completion tools into increasingly autonomous coding agents capable of planning, editing, executing, and validating software~\cite{openai2025gpt5,anthropic2025claude4.5,liu2025deepseek,wang2025openhands,yang2024swe,yang2025code,huang2024opencoder}. This progress has shifted the frontier of automated programming from short-horizon, localized tasks toward a more ambitious goal: end-to-end software construction driven solely by natural-language intent. Achieving this goal requires not only strong code synthesis capabilities, but also sustained long-horizon reasoning, global planning, and cross-file consistency—capabilities that are central to the vision of autonomous software engineering and, more broadly, Artificial General Intelligence (AGI).

Despite this progress, the evaluation landscape has not kept pace with the capabilities being claimed. Most existing benchmarks for coding agents emphasize short-horizon behaviors, such as generating individual functions~\cite{chen2021evaluating, austin2021program}, completing partially specified modules~\cite{liu2025m2rc, liu2023repobench}, or repairing bugs within pre-existing repositories~\cite{jimenez2024swebench, deng2025swe}. While valuable, these settings significantly reduce the demands placed on long-term planning and system-level coherence by providing strong structural priors, limited temporal scope, or frequent human intervention. As a result, \textit{they do not adequately measure whether an agent can sustain coherent decision-making over the hundreds of steps required to design, implement, debug, and finalize a complete software repository.}

Recent repository-level benchmarks partially address this gap, but important limitations remain. Some rely on scaffolded project structures or predefined function signatures~\cite{zhaocommit0}, converting the task into constrained code infilling rather than autonomous construction. Others depend on LLM-based evaluators or qualitative judgments~\cite{staracepaperbench, patwardhan2025gdpval, zhang2025codecriticbench,chou2025autocodebench}, which introduce bias and obscure true functional correctness. Although recent works have expanded evaluation to visual artifacts~\cite{zhang2025artifactsbench} and web agent interactions~\cite{li2025relook}, rigorous evaluation of full-repository construction from natural language remains underexplored. Even benchmarks that use unit tests often assume the presence of an existing codebase~\cite{du2023classeval,jimenez2024swebench}, shifting the challenge toward repair or regression rather than creation. Consequently, \textit{a fundamental question remains unanswered: can current coding agents reliably generate a complete, installable software repository from scratch while maintaining long-horizon coherence?}

\begin{figure}[t]
    \centering
    \includegraphics[width=\linewidth]{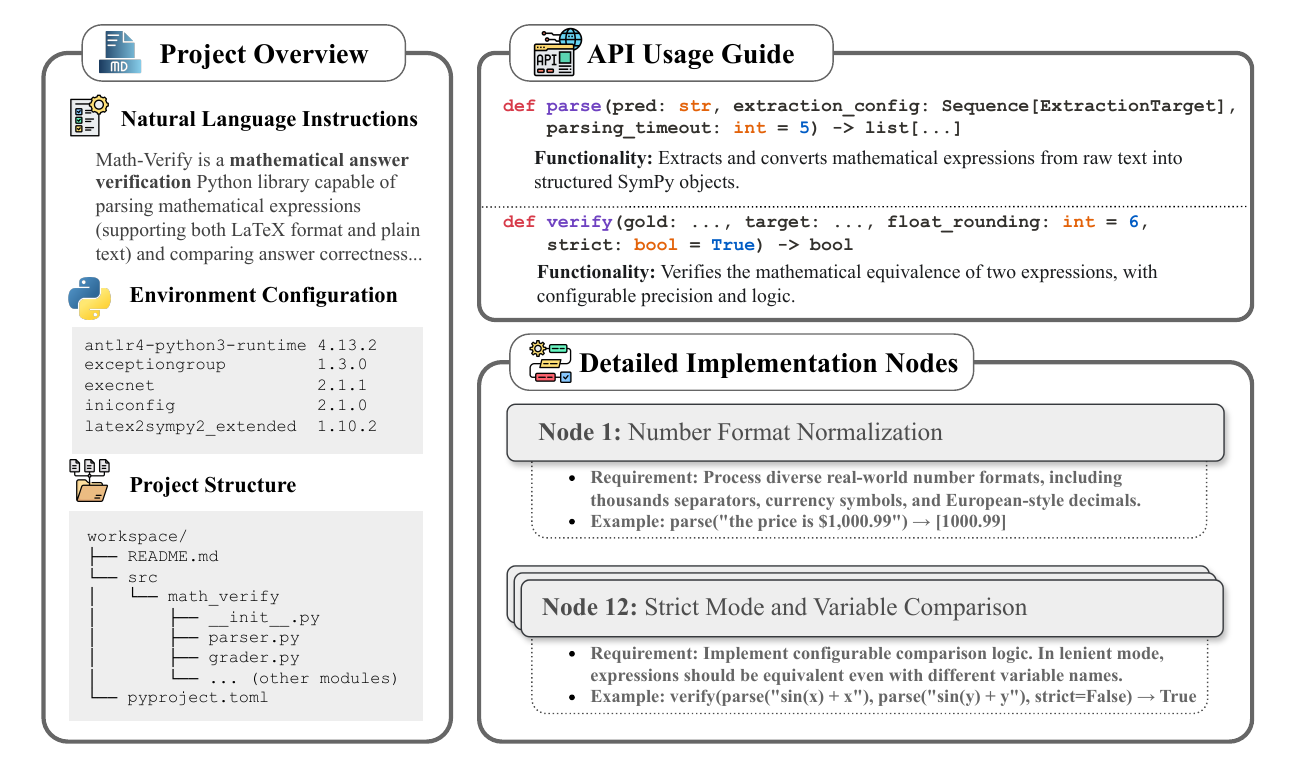}
    \caption{An example NL2Repo task document, illustrating the structured specification (project description, supports, and API usage guide) that agents receive before repository generation.}
    \label{fig:instruction_example}
\end{figure}

To address the above limitations, we introduce \textbf{NL2Repo-Bench}, a benchmark designed to evaluate the long-horizon repository generation capabilities of coding agents. In the NL2Repo-Bench, an agent is provided with a single natural-language requirements document and an empty workspace. From this minimal starting point, the agent must autonomously perform architectural design, dependency management, multi-file implementation, and packaging to produce a fully functional Python library. Crucially, no project scaffolding, source code, or test cases are revealed during development, forcing the agent to reason globally and persistently across the entire construction process. Besides, evaluation in NL2Repo-Bench is strictly execution-based. Each generated repository is verified against the original upstream pytest suite of a real-world open-source project, executed within a controlled environment. This design ensures an objective and binary notion of correctness grounded in real software behavior, rather than subjective judgments or proxy metrics. Moreover, the benchmark comprises 104 tasks drawn from diverse application domains and varying complexity levels, with input documents averaging nearly 19k tokens, reflecting the scale and ambiguity of realistic software specifications.

Through extensive experiments on state-of-the-art (SOTA) open- and closed-source models~\cite{anthropic2025claude4,anthropic2025claude4.5,GoogleCloud2025Gemini3Pro,liu2025deepseek,kimiteam2025kimik2openagentic,openai2025gpt5,glm45agenticreasoningcoding,yang2025qwen3} within multiple agent frameworks~\cite{cursor,claude_code}, we find that long-horizon repository generation remains a major unsolved challenge. Even the strongest agents achieve average test pass rates below 40\% and rarely succeed in fully reproducing a repository. Beyond aggregate performance, our analysis reveals systematic long-horizon failure modes, including premature termination due to overconfidence, loss of global architectural consistency, brittle dependency handling, and an inability to persistently execute and verify plans over extended interaction sequences.

By explicitly targeting long-horizon reasoning and execution, NL2Repo-Bench provides a missing evaluation axis for coding agents and complements existing function-level~\cite{chen2021evaluating,austin2021program} and repair-focused benchmarks~\cite{jimenez2024swebench}. We argue that progress on NL2Repo-Bench will require advances beyond larger context windows, including improved agentic planning, robust self-verification loops, and mechanisms for maintaining global coherence over long development trajectories. As such, NL2Repo-Bench serves both as a diagnostic tool for current systems and as a guiding benchmark for future research on autonomous, long-horizon software engineering.

To summarize, our contributions are as follows:
\begin{itemize}
  \item We formalize the NL2Repo task as constructing a software repository from an empty workspace given only a single requirements document, and introduce \bench{}, a strictly verifiable and long-horizon agentic coding benchmark, which requires the generation of a complete, installable Python library that passes the upstream \texttt{pytest} suite.
  \item We release a reverse-engineered, quality-assured corpus of tasks and a standardized evaluation image that isolates environment effects, enabling apples-to-apples comparisons across agents.
  \item We provide baseline results with state-of-the-art coding agents, revealing substantial gaps versus repair/completion settings and highlighting open challenges in architecture, dependency management, and cross-file consistency.
\end{itemize}

%% file: sections/relatedwork.tex
\section{Related Works}
\subsection{Coding Benchmarks for LLMs}
Function-level benchmarks such as HumanEval and MBPP primarily assess localized reasoning for isolated programming tasks, abstracting away repository-level constraints~\cite{chen2021evaluating,austin2021program}. At the repository scale, prior work clusters into three paradigms. 
(i) \textbf{Repair/Regression}: the SWE-bench series~\cite{jimenez2024swe,zan2025multi,miserendinoswe,deng2025swe,xu2025swe} evaluate resolving real issues within existing projects and validating fixes against project test suites. 
(ii) \textbf{Completion/Auto-completion}: RepoBench~\cite{liu2023repobench} measures the capacity to complete incomplete projects by generating missing components within real repositories~\cite{liu2025m2rc}. 
(iii) \textbf{Paper-to-repository Reproduction}: PaperBench~\cite{staracepaperbench} tests whether agents can replicate AI research by constructing a repository and executing experiments based on papers, with performance often judged by LLMs on code and results rather than authoritative upstream tests. In a different direction, Commit0~\cite{zhaocommit0} targets from-scratch library generation but provides project structure and function signatures as strong priors.

These settings differ along critical axes: input priors (Existing Repository vs. Scaffolding/Signatures vs. Single Document), the evaluation judge (LLM Judgment or Regression Tests vs. Authoritative Upstream Tests), and the required output form (edits within a codebase, completion of components, or fully installable packages). Distinct from all of the above, \bench{} focuses on repository-level generation \textbf{from a single natural-language document}, with \textbf{no scaffolding or signatures}, and uses the upstream repository's \textbf{official pytest suite} as the sole evaluator.

\subsection{LLM-Driven Coding Agents}
The rise of LLMs has given birth to coding agents that assist or autonomously perform software development. Beyond IDE-embedded assistance, more autonomous approaches have emerged. SWE-agent introduces an Agent–Computer Interface (ACI) that abstracts file search, navigation, editing, and testing, enabling LLMs to operate on repository-level tasks and achieve strong results on SWE-bench~\cite{yang2024swe}. OpenHands further generalizes this paradigm as a framework for end-to-end autonomous development, allowing agents to plan, code, and validate with minimal human intervention~\cite{wangopenhands}. These systems generally operate \emph{within} existing codebases. By contrast, \bench{} evaluates agents on constructing a complete, installable repository from a single document, highlighting challenges in architecture design and packaging that are orthogonal to edit/repair settings.

\subsection{Coding-Oriented LLMs}
Large-scale code foundation models form the backbone of advances in automated programming and repository-level code generation. Code Llama~\cite{roziere2023codellama} extends Llama 2 with code-oriented pretraining and long-context support (up to 100k tokens), making it suitable for cross-file reasoning. Open-source model families (e.g., StarCoder-style and DeepSeek-style code models) and recent open cookbooks like OpenCoder~\cite{huang2024opencoder} further push instruction-following and repository-aware capabilities. Advanced training strategies, including domain knowledge distillation~\cite{liu2024ddk} and self-curated data synthesis (e.g., Seed-Coder~\cite{zhang2025seed}), along with reasoning-enhanced models like Seed-Thinking~\cite{seed2025seed}, continue to improve code understanding and generation. While these models improve code understanding and generation, most evaluations still emphasize function-level tasks, underscoring the need for systematic assessment tailored to repository-level generation.

%% file: sections/approach.tex
\section{NL2Repo-Bench}

\begin{figure}[t!]
    \centering
    \includegraphics[width=\linewidth]{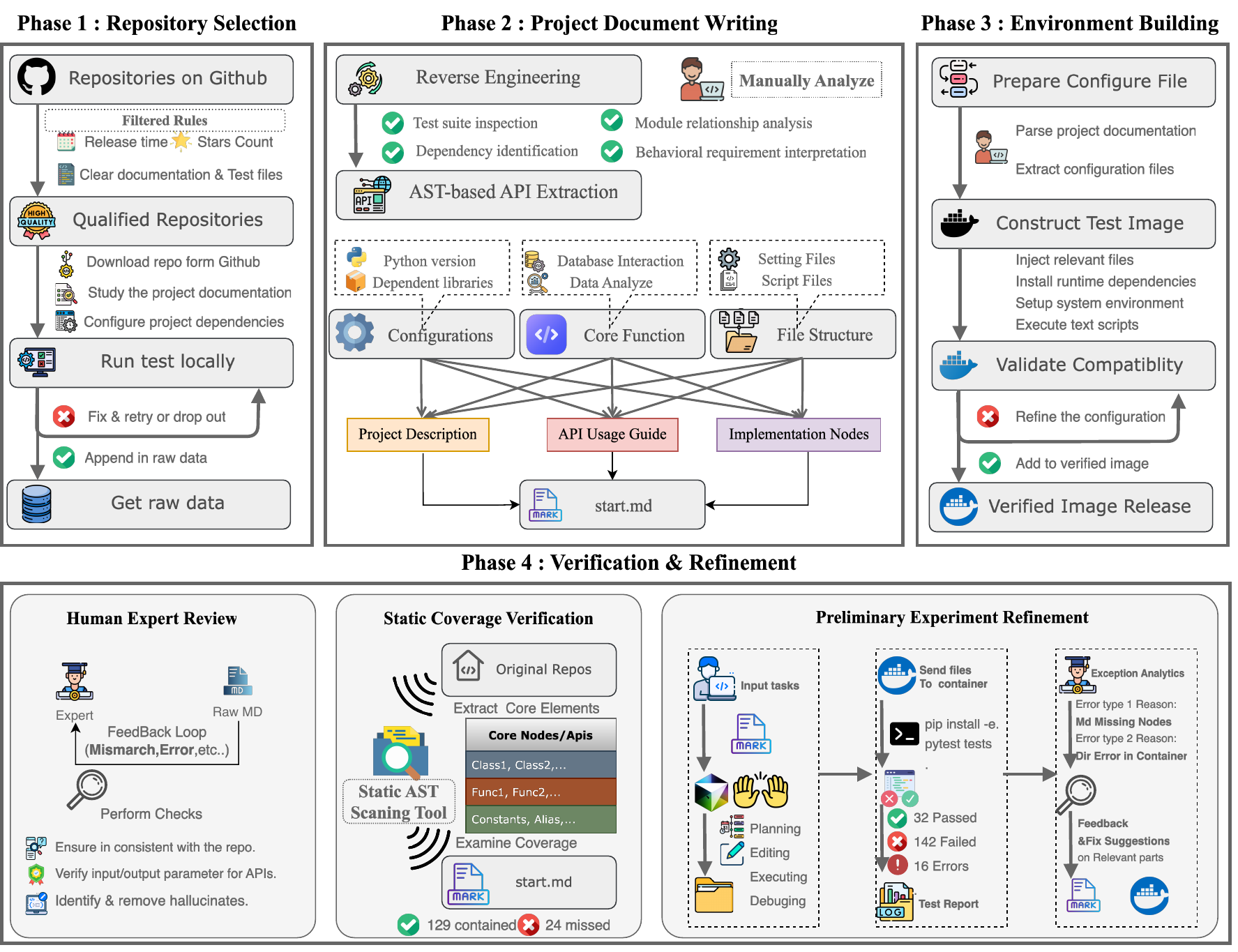}
    \caption{Construction of  \bench{}.}
    \label{fig:frame}
\end{figure}

To evaluate repository-level coding abilities with verifiable ground truth, \bench{} derives tasks from real-world Python libraries characterized by modular architectures and authoritative \texttt{pytest} suites, as shown in Figure~\ref{fig:frame}. Agents receive only a single natural-language specification and must reconstruct the complete repository from scratch, including file structures and functional logic. Correctness is strictly measured by executing the generated code against the original upstream tests. In the following sections, we detail the pipeline for task selection, document generation, and quality assurance (Section~\ref{sec:construction}), followed by the statistical characteristics of the resulting dataset (Section~\ref{sec:statistics}).

\subsection{Benchmark Construction}
\label{sec:construction}

\subsubsection{Repository Selection}
To ensure that each task reflects a realistic and sufficiently challenging repository-level generation scenario, we curate a set of Python open-source libraries from GitHub as the targets to be reproduced. Our selection procedure follows four principled criteria designed to guarantee task complexity, stability, and testability:

\begin{itemize}
\item \textbf{Complexity.} Each repository must contain 300 to 120,000 lines of code. This range excludes trivial projects while avoiding extremely large systems that exceed the context windows of current coding agents, ensuring tasks remain both meaningful and tractable.

\item \textbf{Maturity.} We require each repository to have at least 10 GitHub stars, which serves as a minimal proxy for community adoption, maintenance quality, and functional reliability.

\item \textbf{Completeness and Testability.} A repository must include \texttt{pytest}-based test cases, and its official version must successfully pass all of them. This requirement ensures that the behavioral ground truth is both well-defined and verifiable.

\item \textbf{Recency.} Only repositories that have been created or updated within the past three years are considered, allowing \bench{} to reflect contemporary coding practices and avoid outdated dependency ecosystems.

\end{itemize}

After initial selection, human annotators clone each candidate project and perform an preliminary review of its structure, dependency, and overall organization. Once they develop a sufficient understanding of the repository’s layout and expected behavior, the annotators run the project’s built-in test suite. Only repositories that successfully pass all of their native tests at this stage are deemed qualified and retained as target repositories in our benchmark.

\subsubsection{Project Document Writing}
Upon selecting the target repositories, our annotator construct the input specification through a systematic reverse-engineering process. 
The goal is to translate the entire repository, including implementation files, core functionality, module relationship, and test logic, into a coherent natural-language (NL) document. This document serves as a high-level functional specification that enables a developer (or agent) to fully reproduce the repository's behavior without accessing the source code.
The details of tutorials of our reverse engineering process can be found at Appendix~\ref{app:reverse-engineering tutor}.

To standardize the task formulation, every specification document in \bench{} is structured into four specific sections:
\begin{itemize}
\item \textbf{Project Description:} A high-level overview of the repository's goals, scope, and primary functionality.
\item \textbf{Supports:} Supplementary materials required for development, such as required third-party packages and the expected directory structure.
\item \textbf{API Usage Guide:} Detailed descriptions of the core features to be implemented, including specific requirements for classes, functions, and their expected behaviors.
\item \textbf{Implementation Nodes:} Concrete examples or references for critical APIs (provided where applicable) to ground the agent's implementation planning.
\end{itemize}

Among these sections, the API Usage Guide is particularly crucial. Since \bench{} evaluates generated repositories using the original test suite from the target repository, this section must provide a complete and accurate description of all functional nodes exercised by the tests. Any omission or misaligned API definition would make the task unresolvable for the agent.

To guarantee this completeness, we incorporate an \textbf{AST-assisted} annotation workflow. Annotators begin by running an automated AST scanner over the target repository, which extracts a structured inventory of all functional elements, including class and function names, signatures, and code locations,etc.. Using this inventory as a blueprint, annotators then reconstruct each API entry case-by-case, producing detailed descriptions of expected behaviors, input/output specifications, and semantic constraints. This pipeline ensures that the API Usage Guide is both exhaustive and faithfully aligned with the original repository, enabling reliable long-horizon repository generation tasks.

\subsubsection{Environment Building}
To guarantee deterministic evaluation and isolate code generation quality from environmental variance, we construct a dedicated Docker-based execution environment for each task. The provisioning process follows a strict ground-truth verification protocol. Initially, the environment is configured according to the official dependency documentation of the upstream repository. We then execute the official test suite within this container to establish a baseline. Any execution failures trigger an iterative refinement of the system configuration, such as dependency version pinning or system library adjustments, strictly without modifying the functional source code, until the upstream repository passes all tests.

Furthermore, we implement a strategy of non-functional constraint relaxation within build configurations. We sanitize build manifests (e.g., \texttt{setup.py}) to eliminate rigid constraints that are irrelevant to functional logic. For instance, mandatory existence checks for auxiliary documentation (such as \texttt{README.md} or license files) are either removed or satisfied via the pre-creation of synthetic artifacts. This ensures that agents are evaluated solely on their ability to generate functional software, rather than their adherence to rigid, non-code build prerequisites.

\subsubsection{Verification and Refinement}
To ensure the quality and reliability of the constructed tasks, we employ a multi-stage validation pipeline that integrates both human and automated verification. Our quality assurance framework consists of the following steps:

\begin{itemize}
\item \textbf{Human Expert Review:} Professional Python experts manually verify the fidelity of the source content, the correctness of function signatures(including names, input/output parameters, etc..), and the elimination of hallucinated information, guaranteeing that the document faithfully corresponds to the original codebase.

\item \textbf{Static Coverage Verification:} To ensure that the specification document is sufficiently comprehensive, we perform a static coverage analysis over the target repository. We parse the repository using an AST-based tool to extract all core API definitions, including function names, class structures, and method signatures, and verify that each of these APIs is explicitly documented in the specification with accurate signatures and semantics. This ensures that the specification contains all information necessary for models to implement the full repository from scratch.

\item \textbf{Preliminary Experiment Refinement:} We validate the feasibility of each task by running SOTA coding agents using only the provided specification document. The resulting workspaces are packaged and evaluated within the containerized environments constructed in the previous stage.

To further ensure the reliability of the benchmark design, a senior Python engineer conducts a systematic analysis of all failing test cases and exceptions reported during the test stage. By cross-examining these failures against the task documentation and the corresponding execution environment, the engineer identifies which errors stem from latent issues or ambiguities in the specification or environment configuration, rather than from the model’s capabilities. Based on this diagnosis, we refine the task documents and environments accordingly, ensuring that all benchmark failures are attributable to the model’s reasoning and implementation abilities rather than artifacts of task or environment design. This helps uncover deep semantic ambiguities in the documentation or edge cases in the evaluation environment.
\end{itemize}

A task is recognized as valid only upon passing all verification stages. Any task that fails a check undergoes iterative refinement and re-evaluation until full compliance is achieved.

\subsection{Benchmark Statistics}
\label{sec:statistics}
Following the rigorous selection, construction, and validation pipeline described in Section~\ref{sec:construction}, we assemble the final \bench{} dataset. The benchmark comprises 104 tasks spanning nine distinct categories of Python libraries, as detailed in Table~\ref{tab:cate distribution}. \bench{} represents the first evaluation framework designed to assess coding agents on their ability to generate fully-functional Python repositories solely from natural-language descriptions. The average input length of a \bench{} task is approximately 18,800 tokens, a scale that substantially exceeds the input complexity of existing repository-level benchmarks.

Beyond domain diversity, \bench{} covers a wide spectrum of task complexities. We categorize tasks into three difficulty levels, including easy, medium, and hard, based on the original project size and total lines of code. Detailed criteria and statistics for this categorization are provided in Table~\ref{tab:difficulty_distribution}.

\begin{table}[t]
\centering
\caption{Task categories and statistics of the \bench{}.}
\begin{tabular}{l c}
\toprule
\textbf{Category} & \textbf{Count} \\
\midrule
Web Development & 10 \\
Testing & 13 \\
Utility Libraries & 11 \\
Machine Learning & 7 \\
Data Analysis \& Processing & 18 \\
Database Interaction & 7 \\
Networking Tools & 9 \\
Batch File Processing & 5 \\
System Tools & 24 \\
\midrule
Overall & 104 \\
\bottomrule
\end{tabular}
\label{tab:cate distribution}
\end{table}
\begin{table}[h]
\centering
\caption{Difficulty statistics of the NL2Repo benchmark.}
\begin{tabular}{lcc}
\toprule
\textbf{Difficulty Level} & \textbf{LOC Range} & \textbf{\#Tasks} \\
\midrule
Easy      & $\leq$ 1500 LOC         & 26 \\
Medium    & 1500--4000 LOC          & 46 \\
Hard      & $\geq$ 4000 LOC         & 32 \\
\bottomrule
\end{tabular}
\label{tab:difficulty_distribution}
\end{table}

%% file: sections/experiments.tex
\section{Experiments}

\subsection{Experimental Settings}

In our study, we mainly apply \textbf{\bench{}} to evaluate the performance of various models within the \textbf{OpenHands-CodeAct Agent} framework\footnote{For model \textbf{Gemini-3-pro}, we apply \textbf{Cursor-CLI} framework since it suffers from agent-in-a-loop error frequently in Openhands framework.}. To ensure the comprehensiveness of our experimental results, the evaluation encompassed the following models: \textit{open-source models} — DeepSeek-V3.1, DeepSeek-V3.2\cite{liu2025deepseek}, Qwen3-235B-Instruct\footnote{abbr. Qwen3-Instruct},Qwen3-235B-Thinking~\footnote{abbr. Qwen3-thinking or Qwen3-T}~\cite{yang2025qwen3}, Kimi-k2\cite{kimiteam2025kimik2openagentic}, and GLM-4.6\cite{glm45agenticreasoningcoding}; and \textit{closed-source models} — Claude-Sonnet-4, Claude-Sonnet-4.5~\cite{anthropic2025claude4,anthropic2025claude4.5}, Gemini-3-pro~\cite{GoogleCloud2025Gemini3Pro} and GPT-5~\cite{openai2025gpt5}. 

Considering the impact of different end-to-end agent frameworks on model performance, we additionally employed two commercial coding agents: \textbf{Cursor-CLI} and \textbf{Claude Code}. For both agents, we selected \textit{Claude-Sonnet-4.5} as the underlying language model, ensuring a consistent comparison basis across frameworks while isolating the effect of agent design and tool integration on overall task performance.

The experimental setup is defined as follows: for each model or agent, the working environment consists of an empty workspace containing only the task specification document. After receiving a single initialization instruction from the user, the agent completes the entire task autonomously without any further human input. No additional constraints are imposed on the environment, meaning that the agent is free to invoke any tools available within its system. Considering the complexity of the tasks, we do not set a limit on the number of iteration rounds for the agents.

Once the agent finishes all development processes, the resulting workspace is packaged and passed into a testing image, where the original repository's pytest suite is executed. The average test pass rate is used as the task score for evaluating each model or agent.

To improve the accuracy of the evaluation process, we modified the execution of the original repository tests such that all collected test cases were executed even if errors occurred during the \texttt{pytest} collection stage. This design prevents the entire generated repository from being assigned a zero score due to a small number of collection errors.

\subsection{Main Results}

\begin{figure}[t!]
    \centering
    \includegraphics[width=\linewidth]{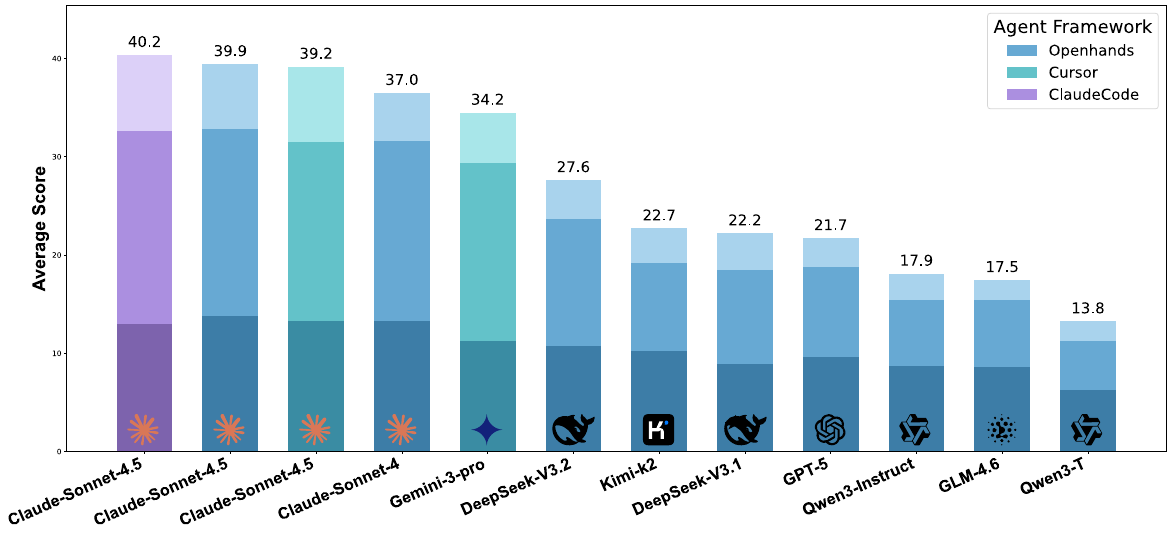}
    \caption{Leaderboard of model performance on NL2Repo-Bench benchmark. Pass rates are evaluated on the official test suite of the target repository. While Claude-Sonnet-4.5 based agents (OpenHands, Claude Code) achieve the highest performance, the best pass rate remains little above 40\%, highlighting the significant challenge of end-to-end repository generation.}
    \label{fig:leaderboard}
\end{figure}

\begin{table}[t!]
\centering
\small
\renewcommand{\arraystretch}{1.1}
\caption{Main Results: Model performance (Pass Rate \%) across difficulty levels. We report the overall pass rate, Pass@1 count, and breakdown by difficulty. Best results are \textbf{bolded}. Unless otherwise specified in parentheses, all models use the \textbf{OpenHands} agent framework.}
\label{tab:main_result}
\resizebox{.8\textwidth}{!}{
\begin{tabular}{lcc|ccc}
\toprule
\multirow{2}{*}{\textbf{Model}} & \textbf{Overall} & \textbf{Pass@1} & \textbf{Easy} & \textbf{Medium} & \textbf{Hard} \\
& \textbf{Score (\%)} & \textbf{(Count)} & \scriptsize{($\le$1.5k LOC)} & \scriptsize{(1.5k-4k LOC)} & \scriptsize{($\ge$4k LOC)} \\
\midrule
\makecell[l]{Claude-Sonnet-4.5 (Claude Code)} & \textbf{40.2} & 3 & 51.8 & \textbf{44.5} & \textbf{25.1} \\
Claude-Sonnet-4.5        & 39.9 & 3 & \textbf{55.3} & 43.0 & 21.4 \\
\makecell[l]{Claude-Sonnet-4.5 (Cursor)} & 39.2 & 4 & 52.9 & 41.4 & 24.8 \\
Claude-Sonnet-4          & 37.0 & \textbf{5} & 53.1 & 41.3 & 16.1 \\
\makecell[l]{Gemini-3-pro (Cursor)} &34.2 &3 &44.9&40.9 &16.8 \\
DeepSeek-V3.2 & 27.6 & 1 & 43.1 & 29.1 & 12.9 \\
Kimi-k2                  & 22.7 & 3 & 40.8 & 20.2 & 11.6 \\
DeepSeek-V3.1            & 22.2 & 1 & 35.7 & 21.6 & 12.1 \\
GPT-5                    & 21.7& 1 & 38.4 & 20.7& 9.6 \\
Qwen3-Instruct                    & 17.9& 1 & 34.7 & 15.2& 8.9\\
GLM-4.6   & 17.5& 2 & 34.4 & 15.5 & 6.5 \\
Qwen3-thinking   & 13.8 & 1 & 25.0 & 11.3 & 6.5 \\

\bottomrule
\end{tabular}
}
\end{table}


\begin{table}[t!]
\centering
\caption{Model performance (Pass Rate \%) across different task categories.}
\resizebox{0.75\textwidth}{!}{
\begin{tabular}{l|ccccc}
\toprule
\textbf{Model} & \textbf{Web Dev} & \textbf{Testing} & \textbf{Utility Libs} & \textbf{ML} & \textbf{Data Proc.} \\
\midrule
Claude-Sonnet-4.5 (Claude Code) & 56.9\% & 30.4\% & 59.8\% & 19.7\% & 36.7\% \\
Claude-Sonnet-4.5 & 37.3\% & 31.9\% & 52.4\% & 16.2\% & 36.8\% \\
Claude-Sonnet-4.5 (Cursor) & 44.0\% & 33.0\% & 53.8\% & 9.4\% & 34.6\% \\
Claude-Sonnet-4 & 48.2\% & 27.5\% & 51.2\% & 8.5\% & 41.3\% \\
Gemini-3-pro (Cursor) & 33.8\% & 28.2\% & 43.9\% & 10.0\% & 26.0\% \\
DeepSeek-V3.2 & 33.6\% & 21.7\% & 44.4\% & 11.1\% & 20.8\% \\
Kimi-k2 & 29.5\% & 18.7\% & 27.3\% & 7.5\% & 24.8\% \\
DeepSeek-V3.1 & 21.2\% & 16.0\% & 30.8\% & 10.5\% & 18.2\% \\
GPT-5 & 26.0\% & 18.6\% & 28.0\% & 7.0\% & 31.6\% \\
Qwen3-Instruct & 18.2\% & 15.2\% & 18.7\% & 8.0\% & 18.9\% \\
GLM-4.6 & 17.7\% & 14.4\% & 29.5\% & 8.4\% & 14.1\% \\
Qwen3-thinking & 15.5\% & 9.1\% & 19.2\% & 7.4\% & 11.6\% \\
\midrule
\textbf{Model} & \textbf{DB Interact.} & \textbf{Network} & \textbf{Batch Ops} & \textbf{Sys Tools} & \textbf{-} \\
\midrule
Claude-Sonnet-4.5 (Claude Code) & 41.3\% & 23.4\% & 43.3\% & 43.6\% & - \\
Claude-Sonnet-4.5 & 44.4\% & 30.5\% & 43.7\% & 50.3\% & - \\
Claude-Sonnet-4.5 (Cursor) & 46.9\% & 25.4\% & 35.0\% & 49.7\% & - \\
Claude-Sonnet-4 & 33.1\% & 23.4\% & 32.0\% & 43.2\% & - \\
Gemini-3-pro (Cursor) & 40.8\% & 28.3\% & 44.0\% & 45.2\% & - \\
DeepSeek-V3.2 & 22.5\% & 16.9\% & 36.4\% & 34.1\% & - \\
Kimi-k2 & 23.8\% & 15.6\% & 37.0\% & 22.3\% & - \\
DeepSeek-V3.1 & 25.0\% & 14.8\% & 33.7\% & 28.0\% & - \\
GPT-5 & 23.4\% & 10.4\% & 18.9\% & 20.2\% & - \\
Qwen3-Instruct & 24.4\% & 14.2\% & 32.6\% & 17.6\% & - \\
GLM-4.6 & 28.9\% & 0.6\% & 20.5\% & 21.0\% & - \\
Qwen3-thinking & 16.6\% & 6.4\% & 16.3\% & 15.8\% & - \\
\bottomrule
\end{tabular}
}
\label{tab:category_performance}
\end{table}

As summarized in Figure~\ref{fig:leaderboard} and Table~\ref{tab:main_result}, several noteworthy observations are as follows:

\textbf{Coding agents remain far from being able to synthesize full repositories.}

As shown in Table~\ref{tab:main_result}, all models achieve an average test pass rate below 40.5\%, and nearly half of them fall below the 20\% threshold. Across the entire set of 104 tasks, the strongest model fully passes the official \texttt{pytest} suite for only five repositories in a single run (Pass@1). These results indicate that current LLMs and coding agents still lack the robustness, long-horizon planning ability, and cross-file consistency required to generate a complete repository from scratch. Even the best-performing systems struggle to construct end-to-end runnable software purely from natural-language specifications, underscoring the substantial gap that remains in achieving reliable repository-level synthesis.

\textbf{Claude outperforms other models.} Across all evaluated models, the Claude series (including Claude-Sonnet-4 and 4.5) demonstrates a clear performance advantage over the others. This superiority is closely tied to its extremely large context window (up to 1M tokens, compared with the $\sim$256K-token limits of most other models) and its ability to sustain long interaction traces without losing track of prior decisions. Given the substantial size of both the input text and the code that must be generated, models with larger context capacities naturally possess a significant advantage, a pattern we analyze in more detail in Section~\ref{sec:context}.

\textbf{GPT-5 underperforms expectations.} Despite GPT-5’s strong performance on other coding-related tasks, its results on NL2Repo-Bench are comparatively weaker. Our case studies suggest a potential explanation: GPT-5 tends to halt prematurely before completing the repository, and waits for additional user input. This observation aligns with our quantitative analysis as well: GPT-5 exhibits the lowest average number of interaction turns, being the only model with fewer than 100 turns, and its turn count is substantially lower than that of other models, as shown in Table~\ref{tab:turns_stats}. These behaviors suggest that GPT-5 is better aligned with human-in-the-loop assistance than with fully autonomous repository generation.

\textbf{Performance degrades with repository complexity.}
To further examine this trend, we compute model scores across tasks of different difficulty levels. A clear pattern emerges: as task difficulty increases, the performance of nearly all models declines substantially. This monotonic degradation confirms that the NL2Repo-Bench difficulty hierarchy effectively captures the intrinsic complexity of real-world, repository-level software development. Moreover, the sharp performance drop on harder tasks highlights a fundamental limitation of current LLM-based coding agents, as they remain far from mastering long-horizon reasoning, multi-module coordination, and dependency-rich engineering workflows that are essential for constructing coherent, large-scale codebases.

\textbf{Performance varies substantially across task categories.}
Table~\ref{tab:category_performance} further breaks down results by library category, revealing that models like Claude-Sonnet-4.5 are particularly strong on system tools and database interaction tasks, while all models struggle more on machine learning and networking tasks. This pattern suggests that current agents handle infrastructure-style repositories with clearer packaging and dependency structure better than domains that require reproducing complex algorithmic pipelines or protocol-heavy logic.

\textbf{Same baseline model behaves similarly across different frameworks.} From Table~\ref{tab:main_result}, we observe that Claude-Sonnet-4.5 exhibits less than a 1\% performance variation across the three agent frameworks, a difference that is negligible given the complexity of NL2Repo-Bench, where each task requires constructing a full repository and passing multiple test cases. In contrast, inter-model performance gaps are substantially larger, indicating that the dominant factor driving NL2Repo-Bench outcomes is the intrinsic capability of the underlying LLM, rather than the specific framework used to orchestrate tool interactions. This result suggests that NL2Repo-Bench behaves primarily as a model-centric benchmark, where improvements to base-model reasoning and generation ability yield far greater gains than modifications to agent-level strategies.

\subsection{Analysis and Discussions}

\subsubsection{Tools Used During Development}
\begin{figure}[t!]
    \centering
    \includegraphics[width=\linewidth]{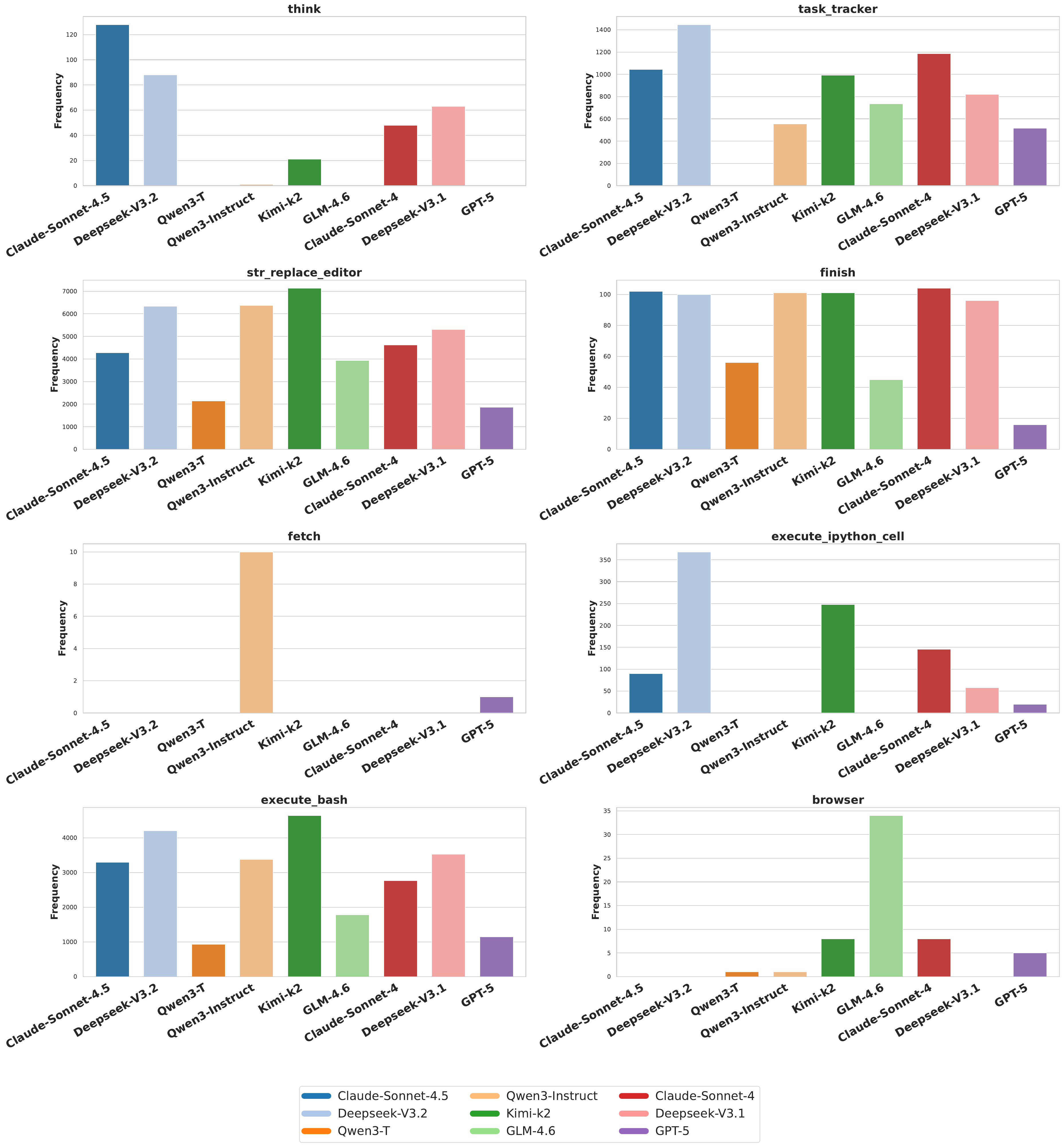}
    \caption{The total times using different tools for models in NL2Repo-Bench Tasks.}
    \label{fig:alltools}
\end{figure}

We analyze the distribution of tool calls across all models within the OpenHands framework, as illustrated in Figure~\ref{fig:alltools}.
The most frequently used tools during the development process are: \texttt{execute\_bash}, \texttt{str\_replace\_editor}, and \texttt{task\_tracker}, which are primarily used for file management \& testing, code editing, and planning, respectively.


We computed the correlation coefficients between the total number of tool invocations and model scores for all models, as shown in Table~\ref{tab:corr}. Among the three most frequently used tools, \texttt{task\_tracker} exhibits the strongest correlation with model performance (0.711), underscoring the central role of effective task planning in constructing large-scale code repositories. In contrast, although the \texttt{think} tool shows a high raw correlation, its low average invocation frequency suggests that this metric may not robustly reflect a causal relationship with overall performance in this specific setting.

\begin{table}[t!]
\centering
\caption{Correlation between tool invocation frequency and model performance.}
\resizebox{0.6\textwidth}{!}{
\begin{tabular}{lcc}
\toprule
\textbf{Tool} & \textbf{Correlation} & \textbf{\#Models Not Using This Tool} \\
\midrule
\texttt{think}                 &  0.816& 4\\
\texttt{task\_tracker}         &  0.711& 1\\
\texttt{finish}                &  0.512& 0  \\
\texttt{execute\_bash}         &  0.371& 0 \\
\texttt{browser}               & -0.264& 3 \\
\texttt{execute\_ipython\_cell} & 0.402& 3\\
\texttt{fetch}                 & -0.1917 & 7 \\
\texttt{str\_replace\_editor}  &  0.161& 0 \\
\bottomrule
\end{tabular}
}
\label{tab:corr}
\end{table}

\begin{table}[t!]
\centering
\caption{Interaction turns statistics and model performance.}
\resizebox{0.5\textwidth}{!}{
\begin{tabular}{lccccc}
\toprule
\textbf{Model} & \textbf{Avg.} & \textbf{Std.} & \textbf{Max} & \textbf{Score} & \textbf{Turns/} \\
& \textbf{Turns} & \textbf{Dev.} & \textbf{Turns} &(\%)& \textbf{Score} \\
\midrule
Claude-Sonnet-4.5          & 181.6 & 64.1 & 394 & 39.9 & 455.1  \\
Claude-Sonnet-4          & 180.7 & 80.3  & 416 & 37.0 & 505.4  \\
Kimi-k2           & 275.1 & 164.6 & 878 & 22.7 & 1211.9 \\
DeepSeek-V3.1     & 202.3 & 170.5 & 990 & 22.2 & 919.6  \\
DeepSeek-V3.2     & 254.3 & 119.9 & 822 & 22.2 & 1145.5 \\
GPT-5             & 78.4  & 29.1  & 165 & 21.7 & 361.2 \\
Qwen3-Instruct    & 212.9 & 166.0 & 940 & 17.9 & 1225.4 \\
GLM-4.6           & 138.6 & 110.7 & 533 & 17.5 & 825.0 \\
Qwen3-Thinking    & 70.2 & 40.9 & 246 & 13.8 & 508.7 \\
\bottomrule
\end{tabular}
}
\label{tab:turns_stats}
\end{table}

\subsubsection{Interaction Turns and Model Performance}

A critical factor distinguishing model performance on \bench{} is the number of interaction turns required to complete repository generation. As shown in Table~\ref{tab:turns_stats}, different models exhibit markedly different interaction patterns.

\textbf{The Dominance of Premature and Incomplete Task Completion in GPT-5.}
A striking finding is GPT-5's significantly lower turn count—averaging only 78.4 turns, which is merely 42\% of Claude-4.5's 181.6 turns and the lowest among all evaluated models. Despite this brevity, GPT-5 achieves a moderate score of 0.217, suggesting high per-turn quality but insufficient task completion. Our case analysis reveals that GPT-5 frequently halts prematurely and awaits user confirmation, with 13.4\% of tasks exhibiting early termination behavior. This pattern indicates that GPT-5's design prioritizes human-in-the-loop collaboration over fully autonomous task completion, misaligning with NL2Repo's requirement for end-to-end repository generation without human intervention.

\textbf{Efficiency vs. Persistence Trade-off.}
Apart from GPT-5, Claude4.5 achieves the best balance between efficiency and performance (turns/score ratio of 455.1), completing tasks with moderate turn counts while maintaining the highest test pass rate. In contrast, DeepSeek-V3.2 employs more turns (254.3) within openhands framework but with mixed results—DeepSeek-V3.2 achieves the third-best score (27.6\%), but with a high cost. This suggests that simply increasing interaction attempts does not guarantee success; the quality of planning and execution strategy matters more.

\subsubsection{The Early Termination and Non-Finish Problem}

Failure to complete the repository structure is a primary cause of low pass rates. We categorize incomplete tasks into two distinct behaviors: \textbf{Early Termination} and \textbf{Non-Finish}.

\begin{itemize}
    \item \textbf{Early Termination (Overconfidence):} The agent explicitly invokes the \texttt{finish} action, claiming the task is done, but does so prematurely (defined here as fewer than 100 interaction turns). This typically indicates a ``false positive'' estimation of progress.
    \item \textbf{Non-Finish (Passive Failure):} The agent never invokes the \texttt{finish} tool. Instead, the session ends because the agent halts to await user input (e.g., asking for clarification or confirmation) or reaches a system timeout. This behavior reflects a lack of agency or an inability to proceed autonomously.
\end{itemize}

\begin{figure}[htbp]
    \centering
    \includegraphics[width=0.9\linewidth]{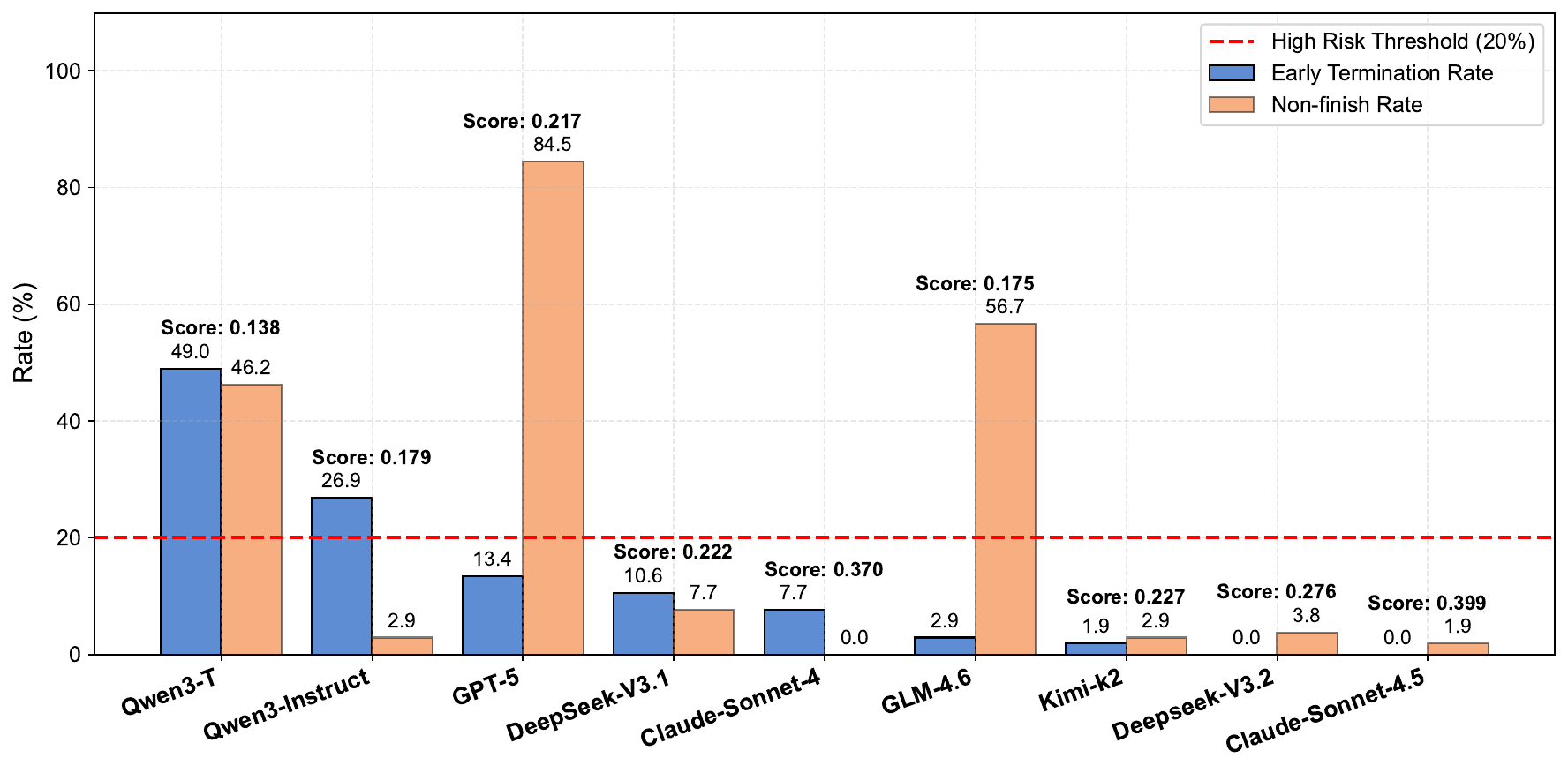}
    \caption{Early termination rates across models. Thinking models (e.g., Qwen3-T) exhibit an alarmingly high early stop rate of 49.0\% and non-finish rate of 46.2\%, leading to unsatisfactory scores.}
    \label{fig:early_stop}
\end{figure}

Figure~\ref{fig:early_stop} quantifies these phenomena. The ``Non-Finish Rate'' is calculated as the percentage of tasks where the agent never invoked the \texttt{finish} tool out of the total 104 tasks, while the ``Early-Stop-Rate'' refers to the percentage of tasks ending within 100 iteration rounds by invoking \texttt{finish} tool.

\textbf{The Thinking Model Paradox.}
Qwen3-Thinking terminates early in 49.0\% of tasks. This high rate suggests that the model's internal reasoning mechanism leads to premature confidence. Rather than verifying implementation details against the requirements, the model ``thinks'' it has completed the work, resulting in incomplete repositories. We hypothesize that this behavior stems from a ``hallucination of verification'' within the model's internal monologue. The thinking process acts as a self-reinforcing echo chamber where the model convinces itself of the correctness of its code through reasoning chains, effectively bypassing the need for actual execution and testing. This ``internal success'' masks the external failure, leading to the observed premature termination.

\textbf{The "Human-in-the-Loop" Dependency of GPT-5.}
A striking finding is the dichotomy between GPT-5's low Early Stop rate (13.4\%) and its massive Non-Finish rate (84.5\%). Unlike thinking models that confidently fail, GPT-5 tends to halt and wait for user guidance (e.g., "I have set up the structure, should I proceed?"). In our fully autonomous benchmark, where no human feedback is provided, this conservative strategy is fatal. It indicates that GPT-5 is aligned more towards collaborative assistance than autonomous execution.

\textbf{Persistent Models.}
Claude-Sonnet-4 demonstrates the most robust agentic behavior, with a 1.9\% Non-Finish rate and a 0 Early Stop rate. It consistently drives the development process to a conclusion, which correlates with its superior performance.

\subsubsection{Tool Usage Patterns and Efficiency}

\begin{figure}[t!]
    \centering
    \includegraphics[width=0.8\linewidth]{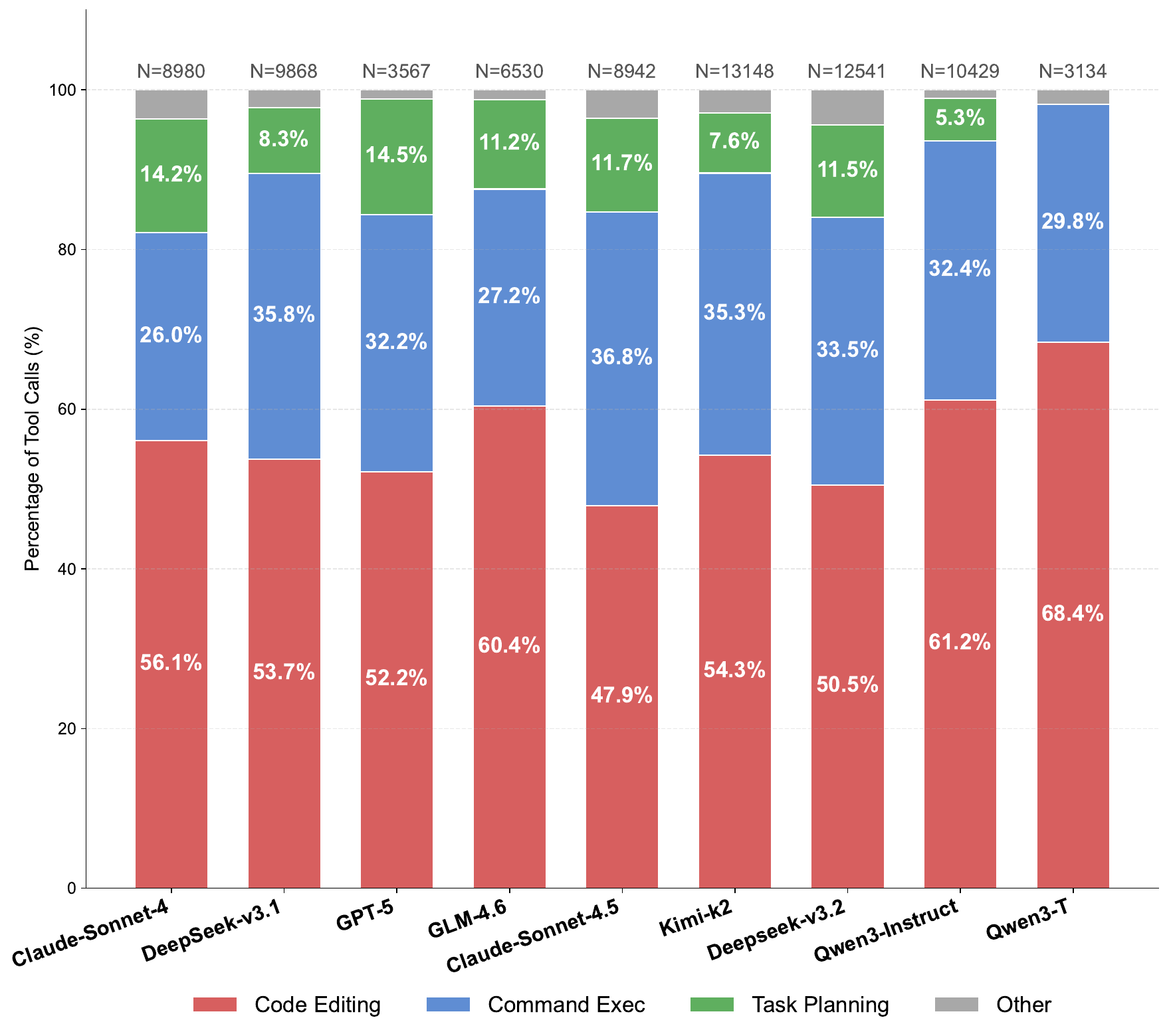}
    \caption{Tool usage distribution across models. Most models allocate 48-62\% of calls to code editing, 26-37\% to command execution, but differ significantly in task planning usage.}
    \label{fig:tool_usage}
\end{figure}

We analyze tool invocation patterns to understand how different models approach repository generation. Table~\ref{tab:tool_usage} and Figure~\ref{fig:tool_usage} present tool usage statistics.

\begin{table}[h]
\caption{Tool usage statistics across models (104 tasks).}
\centering
\resizebox{0.99\textwidth}{!}{%
\small
\begin{tabular}{lcccccclll}
\toprule
\textbf{Tool} & \textbf{Claude} & \textbf{DeepSeek} & \textbf{GPT-5} & \textbf{GLM} & \textbf{Deepseek}& \textbf{Qwen3}& \textbf{Qwen3}& \textbf{Kimi}&\textbf{Claude}\\
& \textbf{-Sonnet-4} & \textbf{-v3.1} & & \textbf{-4.6} &\textbf{-V3.2}&  \textbf{-Instruct}& \textbf{-T}& \textbf{-k2}&\textbf{-Sonnet-4.5}\\
\midrule
str\_replace\_editor & 4623& 5304& 1861& 3955& 6333& 6379& 2143& 7134&4284\\
execute\_bash       & 2766& 3528& 1148& 1781& 4205& 3382& 934& 4642&3293\\
task\_tracker       & 1185& 819& 516& 735& 1447& 554& 0& 994&1045\\
Other               & 306& 217& 42& 79& 556& 114& 57& 378&320\\
\midrule
\textbf{Total}      & 8880& 9868& 3567& 6530& 12541 & 10429& 3134& 13148&8942\\
\textbf{Avg/Task}   & 85.38& 94.88& 34.30& 62.79& 120.59 & 100.28& 30.13& 126.42&85.98\\
\bottomrule
\end{tabular}
}
\label{tab:tool_usage}
\end{table}

\textbf{Code Editing vs. Execution Ratio.}
Most models allocate the majority of tool calls to code editing (str\_replace\_editor), ranging from 48-62\% of total calls. The edit-to-execution ratios reveal different development strategies: Claude-Sonnet-4.5 (1.67:1) and GPT-5 (1.62:1)maintain similar ratios, while DeepSeek-V3.2 uses a lower ratio (1.51:1), suggesting more frequent testing and validation cycles.

\textbf{Task Planning and Organization.}
A notable differentiator is task\_tracker usage, which reflects systematic planning capabilities. GPT-5 devotes 14.5\% of calls to task tracking—the highest among all models—followed by Claude-Sonnet-4 (14.2\%). In stark contrast, Qwen3-Thinking allocates 0 to planning, relying instead on its internal reasoning mechanism. This disparity correlates with performance: models with higher planning tool usage (Claude, GPT-5) achieve better scores, while those neglecting explicit planning (Qwen3-Thinking) suffer from premature termination and incomplete implementations.

\begin{figure}[t!]
    \centering
    \includegraphics[width=0.8\linewidth]{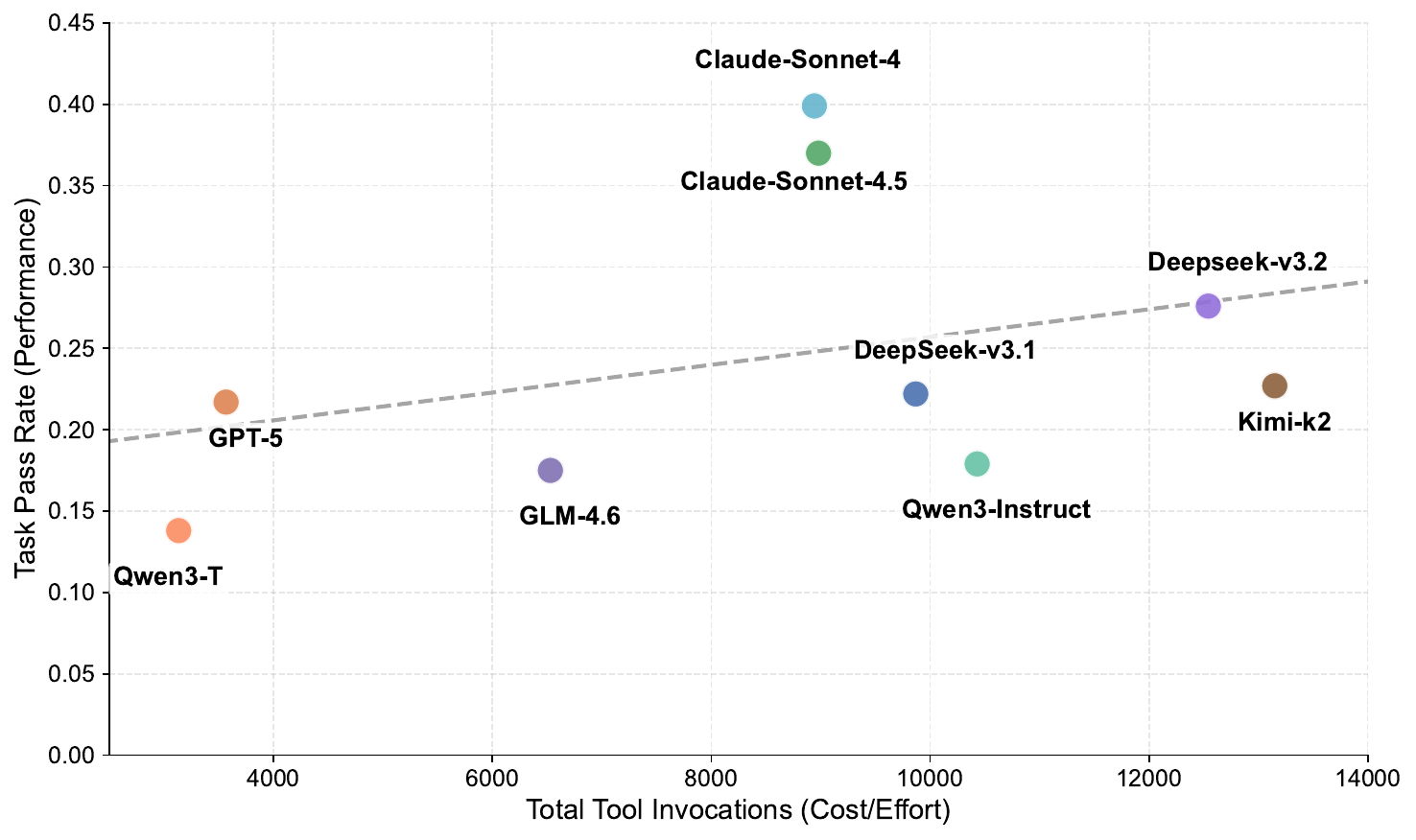}
    \caption{Tool call efficiency: quantity vs. quality. The dashed line represents all model's average performance. GPT-5 and Claude-series operates above this line, while most other models fall below it.}
    \label{fig:efficiency}
\end{figure}

\textbf{Quantity vs. Quality Trade-off.}
Figure~\ref{fig:efficiency} plots total tool calls against final scores, revealing a non-monotonic relationship. GPT-5 achieves a moderate score (0.217) with the fewest calls (3567), demonstrating high per-call quality. Kimi-k2 makes the most calls (13148), but with divergent outcomes, indicating that the sheer quantity of attempts cannot compensate for poor strategy. Claude-4.5-Sonnet's 8942 calls yield the best score, establishing an efficiency baseline.

\subsubsection{Impact of Context Window Size}
\label{sec:context}

\begin{figure}[t!]
    \centering
    \includegraphics[width=0.8\linewidth]{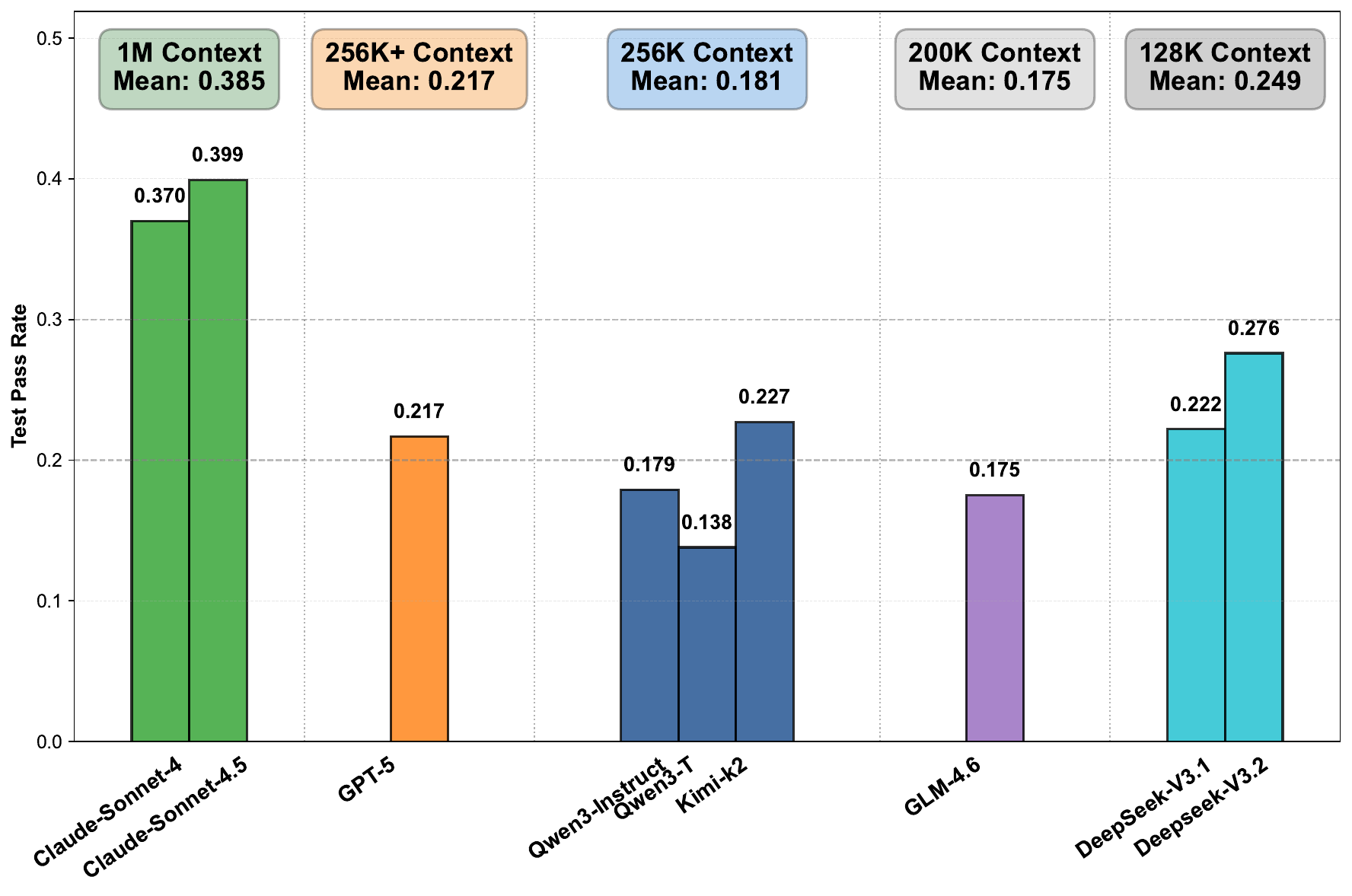}
    \caption{Impact of context window size on performance. The 1M context model (Claude) substantially outperforms 256K models. GPT-5 (256K+) underperforms despite a larger context, suggesting context size is necessary but not sufficient.}
    \label{fig:context_impact}
\end{figure}

Context window capacity emerges as a crucial factor in NL2Repo-Bench performance. Models can be broadly categorized by their context capabilities: those with ultra-long context ($\ge$ 1M tokens, e.g., Claude series, Gemini-3-pro) and those with standard context (e.g., DeepSeek, GPT-5, Qwen). Figure~\ref{fig:context_impact} visualizes performance differences.

\textbf{The Long-Context Advantage.}
The top tier of the leaderboard is exclusively occupied by long-context models. Claude-Sonnet-4.5 (40.2\% on Claude Code) and Gemini-3-pro (34.2\%) substantially outperform the standard-context cohort. This advantage stems from NL2Repo's demanding context requirements: average task documents contain 18,800 tokens, generated code can span 10,000-50,000 tokens, and 180 average interaction turns accumulate approximately 90,000 tokens of conversation history. While 128K capacity theoretically suffices for snapshots, the 1M+ window provides critical headroom for maintaining full context throughout extended development sessions, enabling better cross-file consistency and architectural coherence.

\textbf{Context Size Is Necessary But Not Sufficient.}
However, a large context window does not guarantee superior performance. For instance, despite supporting long contexts, Kimi-k2 achieves a pass rate of only 22.7\%, lagging behind DeepSeek-V3.2 (27.6\%) which operates within a 128K limit. Similarly, GPT-5 (21.7\%) underperforms compared to the top-tier models and even open-source baselines. This suggests that while context capacity provides the \textit{potential} for maintaining global coherence, the model's underlying reasoning capability and agentic behavior (e.g., persistence vs. early termination) are equally critical determinants of success.

\textbf{Context Amplifies Planning Effectiveness.}
While explicit planning (via \texttt{task\_tracker}) is a strategy employed by several models regardless of their context size—DeepSeek-V3.2 (11.5\%), GPT-5 (14.5\%), and Claude-Sonnet-4.5 (11.7\%) all exhibit high usage—the outcomes differ significantly. Standard-context models like DeepSeek utilize planning tools frequently but achieve lower success rates (27.6\%), suggesting that without a massive context window to retain the full history of the plan and its execution states, the effectiveness of planning diminishes over long horizons. In contrast, Claude's 1M+ window allows it to maintain a persistent and coherent view of the task lifecycle, maximizing the utility of its planning actions.

\subsubsection{Agentic Workflow Patterns}

Beyond aggregate tool usage, the \textit{sequence} of actions reveals the agent's underlying reasoning strategy. We analyzed the transition probabilities between consecutive tool calls to identify distinct workflow patterns.

\textbf{The "Edit-Test" Loop.}
High-performing models like Claude-Sonnet-4.5 exhibit a strong cyclic pattern between \texttt{str\_replace\_editor} and \texttt{execute\_bash} (specifically \texttt{pytest}). This ``Edit-Test'' loop indicates a Test-Driven Development (TDD) or rapid feedback strategy, where the agent verifies changes immediately after implementation.

\textbf{The "Navigation" Trap.}
In contrast, lower-performing models show high transition probabilities between \texttt{execute\_bash} (ls/cd) and \texttt{read\_file}, often without intervening edit actions. This ``Navigation Loop'' suggests the agent is struggling to locate relevant files or build a mental map of the repository, leading to wasted context and interaction turns.

\textbf{Blind Editing.}
We also observed a ``Blind Editing'' pattern in some models, characterized by consecutive \texttt{str\_replace\_editor} calls without intermediate testing. This often leads to accumulated errors that are difficult to debug later in the session.

\subsubsection{Failure Taxonomy}

To understand \textit{why} agents fail, we categorized the error types encountered during the evaluation of generated repositories. Unlike function-level benchmarks where \texttt{AssertionError} (logic errors) dominates, NL2Repo-Bench reveals a more complex failure landscape.

\textbf{Environment and Dependency Issues.}
\sloppy
A significant portion of failures  stem from \texttt{ImportError} or \texttt{ModuleNotFound} exceptions. This highlights a key challenge in repository-level generation: agents often struggle to correctly structure the package (e.g., missing \texttt{\_\_init\_\_.py}) or manage internal dependencies between modules, resulting in code that is logically correct but structurally broken.

\textbf{Test Suite Alignment.}
Another common failure mode is the mismatch between the agent's implementation and the official test suite's expectations (e.g., function signatures or class attributes). This suggests that while agents can follow the natural language instructions, they may miss subtle constraints required by the pre-existing tests.

\subsection{Ablation Studies}
We conduct two ablation studies to disentangle how specific evaluation conditions influence model performance on NL2Repo. These analyses clarify whether observed failures arise from benchmark constraints or from fundamental limitations in current coding agents.

\subsubsection{Impact of interaction round limits}
\begin{figure}[htbp]
    \centering
    \includegraphics[width=0.7\linewidth]{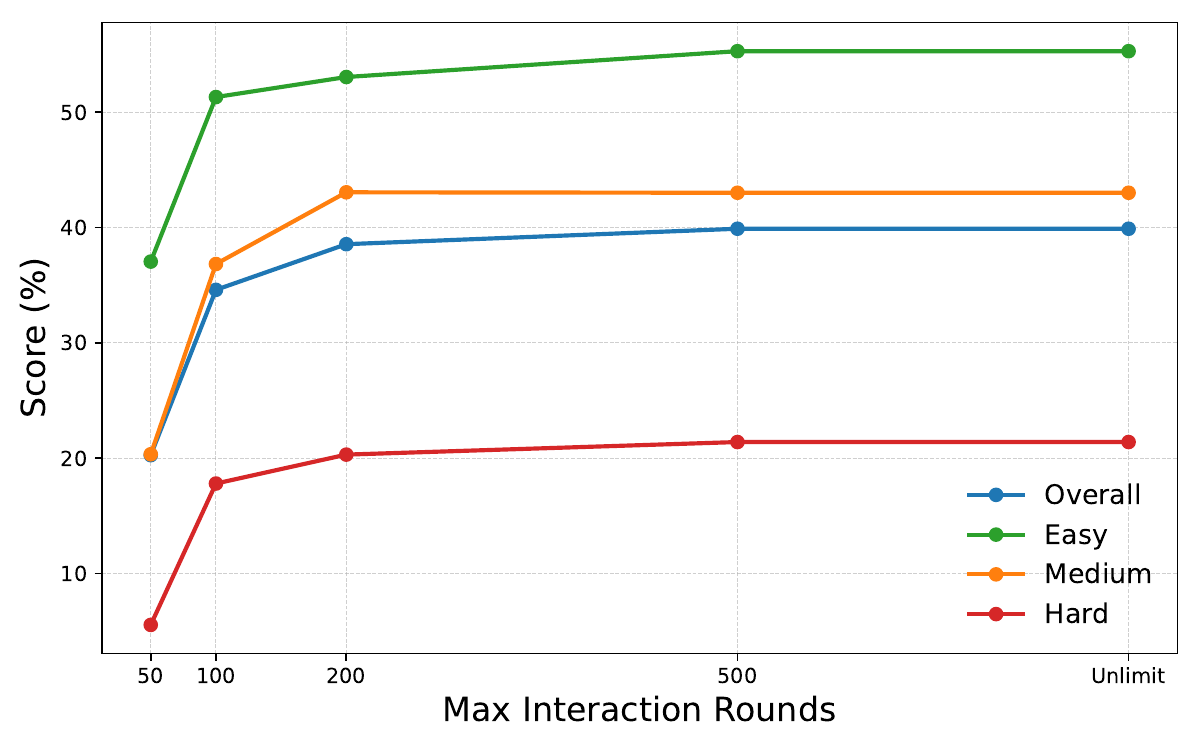}
    \caption{The effect of limitations on iteration rounds on the performance of Claude-Sonnet-4.5 on different difficulty level tasks.}
    \label{fig:roundlimit}
\end{figure}

The main experiments permit unlimited interaction rounds, allowing agents to iteratively refine the repository without budget constraints. To assess whether high scores depend on extensive trial-and-error rather than genuine reasoning, we re-evaluate each model under varying maximum round limits. This ablation quantifies the degree to which agents rely on long iterative loops and examines their effectiveness under more realistic, bounded interaction settings. As shown in Fig.~\ref{fig:roundlimit}, model performance increases steadily across all difficulty levels as the interaction limit is raised from 50 to 200 rounds. Once the limit reaches 200 rounds---which is slightly above Claude-Sonnet-4.5’s average number of interactions under the unrestricted setting---further expanding the maximum round budget yields only marginal improvements. These findings indicate that while interaction budget matters in the low-round regime, model performance quickly saturates once the budget exceeds its natural working range, suggesting that the primary limitations lie in semantic reasoning, architectural planning, and cross-file consistency rather than in the sheer number of allowed interaction steps.

\subsubsection{Impact of revealing all test cases.}
By default, agents only receive the natural-language specification, while the pytest suite remains hidden to mimic real development settings. To estimate an upper bound on performance and determine how much models struggle with implicit requirement inference, we run an additional condition where all test cases are made visible during generation. Comparing the two settings reveals whether performance bottlenecks stem primarily from semantic understanding of the NL document or from deeper challenges in implementing a coherent, multi-file system even with full supervision. 

Results in Table~\ref{tab:unhack} show that exposing the full test suite yields substantial performance gains. For example, the pass rate of Claude-Sonnet-4.5 (in Claude Code framework) increases markedly from 40.2\% to 59.4\%, and the number of fully-passed tasks jumps from 3 to 18. This trend is consistent with the expected benefit of providing test cases, which can guide and constrain both the development process and the resulting implementation. However, it is equally noteworthy that—even under this ``cheating'' scenario where evaluation-phase tests are made available during development—the model’s overall score still does not exceed 60\%. This indicates that, despite the advantages of test-driven development, generating a fully functional, end-to-end runnable repository from scratch remains a substantial challenge for current coding agents, pointing to fundamental limitations in long-horizon coordination and large-scale code synthesis rather than merely missing supervision signals.

\begin{table}[t!]
\centering
\small
\renewcommand{\arraystretch}{1.15}

\caption{Comparison of the performance of Claude-Sonnet-4.5(Claude Code) on whether revealing all test cases. }
\label{tab:unhack}
\resizebox{\linewidth}{!}{
\begin{tabular}{lcc|cc|cc|cc}
\toprule
\textbf{Inputs} &
\multicolumn{2}{c|}{\textbf{Easy}} &
\multicolumn{2}{c|}{\textbf{Medium}} &
\multicolumn{2}{c|}{\textbf{Hard}} &
\multicolumn{2}{c}{\textbf{Overall}} \\
\cmidrule(lr){2-3} \cmidrule(lr){4-5} \cmidrule(lr){6-7} \cmidrule(lr){8-9}
 & \textbf{Avg Score (\%)} & \textbf{Pass@1}
 & \textbf{Avg Score (\%)} & \textbf{Pass@1}
 & \textbf{Avg Score (\%)} & \textbf{Pass@1}
 & \textbf{Avg Score (\%)} & \textbf{Pass@1} \\
\midrule
Document only         & 51.8 & 1& 44.5 & 1 & 25.1 & 1 & 40.2 & 3\\
Document + unittest   & 73.2 & 9 & 67.5 & 7 & 35.6 & 2 & 59.4 & 18 \\
\bottomrule
\end{tabular}
}

\end{table}

\section{Conclusion}

We introduce \textbf{\bench{}}, the first benchmark to rigorously evaluate the capability of LLMs and agents to generate complete, installable Python repositories from scratch, starting from a single natural language document. By separating development from evaluation and using authoritative upstream test suites, we provide a strict measure of autonomous software engineering capability.

Our extensive experiments reveal a substantial gap between current SOTA models and the requirements of repository-level generation. While models like Claude-Sonnet-4.5 show promise with a 39.6\% pass rate, all agents struggle with the complexity of long-horizon planning and cross-file consistency. We identified two critical failure modes: the ``overconfidence'' of thinking models leading to early termination, and the ``collaborative bias'' of models like GPT-5 that fail to proceed autonomously.

NL2Repo-Bench serves as a testbed for the next generation of coding agents. We believe that progress in this domain will require not just larger context windows, but architectural innovations in agentic planning, self-correction loops, and reliable environment management. We release the NL2Repo benchmark, including the dataset, docker environments, and evaluation toolkit, to the research community.

%% file: sections/contribution.tex
\section{Contributions}

\textbf{Leading Authors}($\alpha$-$\beta$ order)\\ 
Jingzhe Ding, Shengda Long, Changxin Pu, Ge Zhang, Huan Zhou  

\textbf{Core Contributors} ($\alpha$-$\beta$ order)\\
Hongwan Gao, Xiang Gao, Chao He, Yue Hou, Fei Hu, Zhaojian Li, Weiran Shi, Zaiyuan Wang, Daoguang Zan, Chenchen Zhang, Xiaoxu Zhang \\

\textbf{Contributors} ($\alpha$-$\beta$ order)\\
Qizhi Chen, Xianfu Cheng, Bo Deng, Qingshui Gu, Kai Hua, Juntao Lin, Pai Liu, Mingchen Li, Xuanguang Pan, Zifan Peng, Yujia Qin, Yong Shan, Zhewen Tan, Weihao Xie, Zihan Wang, Yishuo Yuan, Jiayu Zhang, Enduo Zhao, Yunfei Zhao, He Zhu, Liya Zhu, Chenyang Zou 

\textbf{Sponsor Commitee}($\alpha$-$\beta$ order)\\
Ming Ding, Jianpeng Jiao, Jiaheng Liu, Minghao Liu, Qian Liu, Chongyang Tao, Jian Yang, Tong Yang, Zhaoxiang Zhang \\

\textbf{Corresponding}\\
XinJie Chen 
(\email{chenxinjie.bj@bytedance.com})\\ 
Wenhao Huang (\email{huang.wenhao@bytedance.com})\\ 
Ge Zhang
(\email{zhangge.eli@bytedance.com})\\


%% file: sections/appendix.tex
\section{Difficulty Level in NL2Repo Tasks}
\label{app:difficulty level}
We classify the difficulty of tasks in NL2Repo based on the number of lines of code (LOC) in the original repository. As shown in Table~\ref{tab:difficulty_distribution}, repositories with fewer than 1,500 LOC are categorized as \textbf{Easy}, while those exceeding 4,000 LOC are categorized as \textbf{Hard}. Repositories falling between these thresholds are considered \textbf{Medium}.

\section{Available Tools in OpenHands CodeAct Framework}
\label{app:tools}

In our experiments, agents interact with the environment using the following standardized tools provided by the OpenHands CodeAct framework:

\begin{itemize}
    \item \textbf{execute\_bash}: Executes a bash command in a persistent shell session. This tool is essential for file navigation, package installation, and running test suites.
    \item \textbf{think}: Enables the agent to articulate internal reasoning traces and plan next steps without executing any changes in the environment.
    \item \textbf{finish}: Signals the completion of the current task.
    \item \textbf{browser}: Allows the agent to interact with a web browser using Python code (e.g., for documentation lookup).
    \item \textbf{execute\_ipython\_cell}: Runs a cell of Python code in an interactive IPython environment.
    \item \textbf{task\_tracker}: Provides structured task management capabilities, allowing the agent to view, add, and update the status of development tasks.
    \item \textbf{str\_replace\_editor}: A custom file editing tool designed for viewing, creating, and editing files. It uses strict string matching to ensure precise code modifications.
    \item \textbf{fetch}: Retrieves content from a specified URL and optionally extracts it as markdown.
    \item \textbf{create\_pr / create\_mr}: Tools to simulate the submission of a Pull Request or Merge Request on platforms like GitHub, GitLab, or Bitbucket.
\end{itemize}

\section{Tutorial for NL2Repo-Bench Annotators}

\label{appendix:annotator_tutorial}

This section provides the guidelines and step-by-step workflow used by
annotators to construct task specifications for NL2Repo.
The goal is to ensure that all specifications are
(1) semantically faithful to the original repository,
(2) sufficiently comprehensive for end-to-end repository development,
and (3) consistent across annotators.
The process consists of three major phases:
project selection, repository comprehension and environment validation,
and structured specification writing. The procedures summarized below are
derived from the operational annotation guidelines used in practice.\footnote{Summary based on internal annotation draft.} 

\subsection{Phase 1: Project Selection}
Following the criteria in Section~3.1.1 of the main paper,
annotators first determine whether a candidate GitHub repository is
eligible for inclusion.
This includes checking project maturity, testing completeness, license
compatibility, and the feasibility of isolating the core functionality.

\subsection{Phase 2: Repository Understanding \& Test Validation}

Annotators must obtain a precise and executable understanding of the target repository before writing its specification.

\textbf{(1) Local Setup and Preliminary Analysis.}
\begin{itemize}
    \item Download the repository locally from GitHub.
    \item Conduct an initial review of the project, including its purpose,
          core functionality, directory structure, and external dependencies.
\end{itemize}

\textbf{(2) Environment Construction.}
Annotators create an isolated environment based on the repository’s
documentation (e.g., \texttt{README}, \texttt{requirements.txt},
\texttt{setup.py}, or \texttt{pyproject.toml}).
All relevant dependencies, including possible undocumented ones, must be
installed.

\textbf{(3) Full Test Execution.}
Annotators must run \emph{all} existing test cases in the repository.
A repository is considered valid only if all tests pass.
If failures occur, annotators must diagnose and resolve environment-related issues
(e.g., Python version mismatches, incompatible dependency versions, missing system packages).
Repositories with unresolvable failures are excluded.

\subsection{Phase 3: Task Specification Construction}
\label{app:reverse-engineering tutor}
Each task requires a comprehensive specification consisting of three parts:
a project-level description, support information, and an API-level usage guide.

\subsubsection{Project Description}
Annotators provide a high-level functional summary of the repository,
describing:
\begin{itemize}
    \item the overall purpose and design of the project;
    \item its major components and interaction patterns.
\end{itemize}
Annotators may reference official documentation or use LLM-based tools
to draft an initial summary, but the final text must be manually verified
to ensure factual correctness.

\subsubsection{Support Information}

\textbf{(1) Third-Party Dependencies.}
Annotators list all external libraries needed to run the project,
including explicit dependencies and any additional packages required for the tests to pass.
Version numbers must be preserved where applicable.

\textbf{(2) Repository File Structure.}
A complete directory layout of the target codebase is included
(excluding the testing directory).
All files relevant to the implementation must be listed to ensure that
the model can reconstruct the same structure during development.

\subsubsection{API Usage Guide}

This component provides natural-language documentation for all functional units in the repository.

\textbf{(1) Static Extraction of Functional Nodes.}
Annotators run a static analysis tool to extract:
\begin{itemize}
    \item all classes, functions, and constants;
    \item their names and signatures;
    \item their file locations.
\end{itemize}
Test directories are excluded from the scan.

\textbf{(2) Node-Level Documentation.}
For each functional node, annotators write a standalone description containing:
\begin{itemize}
    \item the name and purpose of the node;
    \item input arguments and return values (with additional explanation if complex);
    \item a functional description aligned strictly with the implementation.
\end{itemize}
Text may be drafted with AI tools for assistance, but annotators must manually verify
that every detail matches the underlying source code exactly.

\textbf{(3) Module Import Instructions.}
Annotators provide examples of how modules and APIs should be imported.
These import paths are derived from the project’s internal import patterns
as observed in the test files (excluding imports from external libraries).

\subsection{Implemention Nodes}
Annotators provide concrete examples or references for critical APIs (provided where applicable) within the target repository to help understand the specific usage of APIs. When completing this part, annotators could refer to the examples in the repository.

\subsection{Quality Requirements}
All descriptions must be:
\begin{itemize}
    \item fully consistent with the repository’s implementation;
    \item complete with respect to all functional nodes;
    \item free of speculative or missing behaviors not grounded in the source code.
\end{itemize}
This ensures that the resulting specification is both comprehensive and
faithful, enabling an agent to implement the entire repository solely from the provided instructions.